\documentclass[12pt,a4paper]{cibb}

\makeatletter
\providecommand{\@ordinalM}[2]{#1}
\makeatother

\usepackage{subfigure,graphicx}
\usepackage{amsmath,amsfonts,latexsym,amssymb,euscript,xr}
\usepackage{booktabs}
\usepackage[nodayofweek]{datetime}
\usepackage{hyperref}
\usepackage{fmtcount}
\usepackage[english]{datenumber}
\usepackage[absolute]{textpos}
\usepackage{multirow}

\usepackage[table]{xcolor}
\usepackage{color,colortbl,tabularx}

\usepackage[english]{babel}
\usepackage[protrusion=true,expansion=true]{microtype}
\usepackage{amsmath,amsfonts,amsthm}
\usepackage{pifont}

\definecolor{LightBlue}{rgb}{0.88,0.9,0.9}

\title{\Large $\ $\\ \bf An unsupervised clustering analysis of breast cancer data derived from electronic health records enhanced through UMAP dimensionality reduction}

\author{\large Davide Chicco$^{*,1,2}$, and Nicoletta~Benvenuto$^{1}$}
\address{\footnotesize $\ $\\$^1$ Dipartimento di Informatica Sistemistica e Comunicazione, Università di Milano-Bicocca, Milan, Italy.  \\
$^2$ Institute of Health Policy Management and Evaluation, University of Toronto, Toronto, Ontario, Canada.
\\
\bigskip
ORCID codes: DC~0000-0001-9655-7142 and NB~0009-0008-9827-8786.
\bigskip
\newline
$^*$corresponding author: davidechicco@davidechicco.it
}

\abstract{\small clustering; DBSCAN; electronic health records; dimensionality reduction; UMAP; breast cancer; cancer. \normalsize
\\[17pt]
{\bf Abstract.} 
Breast cancer is one of the most widespread types of cancer, affecting approximately 8 million women worldwide.
Electronic health records of patients diagnosed with this disease can serve as valuable datasets for computational analyses, enabling the discovery of new insights about the pathology.
Unsupervised clustering, in particular, can identify groups of patients with medically significant features, revealing data trends that might otherwise go unnoticed by medical doctors.
In this study, we first applied the DBSCAN density-based clustering method to three independent datasets derived from electronic medical records of patients with mammary carcinoma.
Subsequently, to enhance our results, we preceded the DBSCAN application with a dimensionality reduction phase using UMAP.
We evaluated our clustering outcomes using three statistical indices (DBCV, DCSI, and DISCO).
Our results confirm the effectiveness of combining UMAP with DBSCAN for clustering data derived from electronic health records, paving the way for the medical interpretation of the patient groups identified by our approach.
}

\begin{document}

\renewcommand{\thefootnote}{}
\footnotetext{\small{Article version: \datedate $\;$ h\currenttime  $\;$ CET}}

\thispagestyle{myheadings}
\pagestyle{myheadings}
\markright{\tt Proceedings of CIBB 2026}

\section{Introduction}
\label{sec:SCIENTIFIC-BACKGROUND}

Breast cancer is the most common cancer worldwide among women, with an estimated 2.3 million new cases diagnosed globally in 2022. 
Each year, approximately 670 thousand deaths are caused by breast cancer worldwide, and it is estimated that about one in twenty women will develop the disease during their lifetime. 
Around 30 thousand new cases of mammary carcinomas affect men every year.

Electronic health records (EHRs) contain valuable patient data and can be used to infer new knowledge through computational methods. 
Analyzing electronic health record data from patients diagnosed with this type of cancer using computational intelligence and statistical methods can be useful for identifying new trends and patterns, as well as generating new insights into the disease.

Clustering, in particular, is a family of unsupervised machine learning methods that groups patients based on similarities in their data, and DBSCAN is a density-based clustering technique well suited for this purpose. Dimensionality reduction, that is the projection of high-dimensional data into a lower-dimensional space while preserving the global and local structure of the original dataset, is known to improve clustering performance~\cite{allaoui2020considerably}.

In this study, we applied DBSCAN clustering to three publicly available datasets of medical records from patients with breast cancer and evaluated the results using common internal clustering metrics (DBCV, DCSI, and DISCO). We then repeated the same procedure, this time preceding DBSCAN with a dimensionality reduction step using UMAP, an efficient projection algorithm.
Our results show that using UMAP with appropriately chosen hyperparameters can improve the clustering performance of DBSCAN on these EHRs datasets.

\section{Data and Methods}
\label{sec:DATA-AND-METHODS}

\paragraph{Datasets}
The datasets analyzed in this study are three (Seoul2016, Qingdao2016, and Heidelberg2019), and all are derived from electronic health records of patients with breast cancer.

The Heidelberg2019~\cite{michel2019prediction} dataset consists of 456 records and 61 attributes. 
The data, obtained from the Heidelberg Breast Cancer Unit for Diagnosis and Treatment of PBC, were collected between 2003 and 2011 and relate to patients diagnosed with breast cancer and treated with NAC (neoadjuvant chemotherapy).
The collected data include information on tumor characteristics, such as histology; ER, PR, and HER2 status; tumor laterality (right/left); KI67 protein levels; and molecular subtype classification (based on the classification introduced at the 2013 Saint~Gallen Consensus Conference). Additional information includes type of therapy (chemotherapy, radiotherapy, endocrine therapy), survival outcomes, time since last follow-up, DMFS (distant metastasis-free survival), and LRRFS (locoregional recurrence-free survival).
The dataset also contains, for each patient, the MD Anderson Prognostic Index (MDAPI) \cite{chen2005breast}, a prognostic score ranging from 0 to 4. 
It is calculated as the sum of four components related to the clinical status of lymph nodes, the size of the residual pathological tumor, the pattern of residual disease, and the presence of lymphovascular space invasion in the surgical specimen.

The Qingdao2016~\cite{jiao2016elevated} dataset contains 400 records, divided into two groups: 200 patients diagnosed with breast cancer (case group) and 200 healthy women (control group).
The data were collected by the Affiliated Hospital of Qingdao University Medical College between August 2012 and December 2013 through blood sampling prior to the start of treatment.
The original study investigated the relationship between serum levels of RBP4 (retinol-binding protein 4) and breast cancer risk. Protein levels were measured using enzyme-linked immunosorbent assay (ELISA) on separate serum samples.
The dataset contains 18 attributes, including patient history (age, age at menarche, menopause status, miscarriages, and body mass index), diagnostic information (ER, PR, metastasis status, metastasis stage, and histopathological subtype), and blood test results (fasting blood glucose, triglycerides, total cholesterol, HDL, LDL, P53, and RBP4 protein levels).
In this study, we used only the data corresponding to the 200 patients diagnosed with breast cancer.

The Seoul2016~\cite{lee2016clinicopathological} dataset consists of data from 104 patients diagnosed with ILRR (isolated locoregional recurrence) who were treated between January 2000 and December 2010 at the Samsung Medical Center.
The dataset contains 51 attributes covering information from both the initial diagnosis (type of surgery, ER, PR, HR, HER2 status, subtype and histological type, tumor stage, and type of therapy) and the ILRR diagnosis, which includes the same set of clinical variables. 
It also includes patient characteristics such as age, survival outcomes, and the occurrence of distant metastases. 
In addition, it records temporal information describing disease progression, such as the time interval between the first surgery and the diagnosis of ILRR, the interval between ILRR diagnosis and the development of distant metastases, and the overall duration of follow-up.

\paragraph{Methods}
We analyzed these datasets through the DBSCAN clustering method, first on each original dataset and then on its low-dimensional representation obtained by UMAP. 
DBSCAN (Density-Based Spatial Clustering of Applications with Noise) \cite{khan2014dbscan} is a non-supervised clustering algorithm, based on density. The main difference between this method and other common clustering methods such as $k$-means, is that the resulting clusters (that are found with no need to specify the number of desired clusters) are created by finding ``core samples'' and then adding neighbors points. 
DBSCAN uses two fundamental parameters: epsilon $\epsilon$ and minimal samples.
Epsilon is a real value and the maximum distance between two points in order to consider them as neighbors.
Minimal samples is a natural integer value, and is the minimum number of points (including the point itself) that composes the neighborhood of a point in order to consider it as a ``core sample''.

The choice of the value of these parameters has an effect on the resulting clusters. For instance, a smaller epsilon will lead to more clusters, while a high value of minimal samples will lead to the creation of bigger clusters.

%

%

Eventually, DBSCAN assigns points to real clusters and outliers to the noise cluster.

We first applied DBSCAN to each of the three datasets, and then decided to perform a dimensionality reduction through UMAP on them before applying UMAP.

UMAP (Uniform Manifold Approximation and Projection) \cite{Healy2024} is a non-linear dimensionality reduction algorithm, that considers three main assumptions on data: data is uniformely distributed on a Riemann manifold, Riemann metric is locally constant, or can be approximated as it a constant, and Riemann manifold is locally connected.
Its main hyperparameters are the number of neighbors (dimension of local neighborhood), and minimal distance (minimum distance between points).

%
Since the main goal of this study is demonstrate that applying a dimensionality reduction on datas could enhance the clustering results, UMAP method has been applied to each dataset before applying DBSCAN clustering method.

We performed a grid search for hypeparameter tuning both for DBSCAN and UMAP.
For DBSCAN, we chose the value of minimal samples in a range of values between 3 and the number of attributes of the dataset plus 1.
We selected  the value of epsilon using the $k$-distances plot, which represents the distances, in ascending order, between a point and its ``nearest neighbors.''

To select the UMAP parameters, we performed several tests with different value ranges of neighbors' number and minimum distance. 
For the Seoul2016 dataset, we tested the neighbors' number 5, 10, 15, and 20, and the minimal distances from 0 to 0.25 with 0.05 step.
For the Qingdao2016 dataset, we tested the neighbors' number from 2 to 20 with 1 step, and the minimal distances from 0.01 to 0.25 with 0.05 step.
For the Heidelberg2019 dataset, we tested the neighbors' number from 2 to 31 with 5 step, and the minimal distances from 0 to 0.51 with 0.05 step.
We selected the UMAP configuration leading to the higher trustworthiness.
Then we applied DBSCAN to the resulting reduced dataset and calculated the DBCV value.
Finally, we chose the combination of UMAP and DBSCAN parameters that yielded the best DBCV results.

We assessed the results through the DBCV~\cite{moulavi2014density}, DCSI~\cite{gauss2023dcsi}, and DISCO~\cite{beer2025disco} indices.
To avoid empty and irrelevant clusters, we set the following constraints: DBCV index at least +0.5 in the $[-1,+1]$ range, cluster sizes at least 10\%, and maximum size of the noise cluster 20\%.
The DBCV index is a density-based version of the Silhouette coefficient and we consider it the most informative internal clustering assessment coefficient available in the scientific literature.

\section{Results}
\label{sec:RESULTS}

\begin{figure}[h]
\vspace{3mm}
 \begin{center}
    (a)~\includegraphics[width=0.45\linewidth]{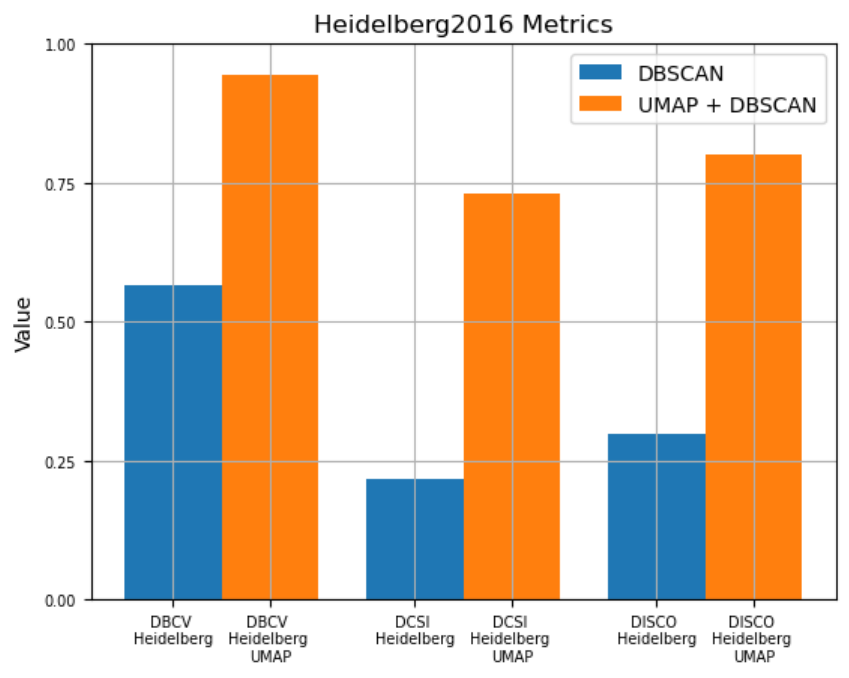}
    (b)~\includegraphics[width=0.45\linewidth]{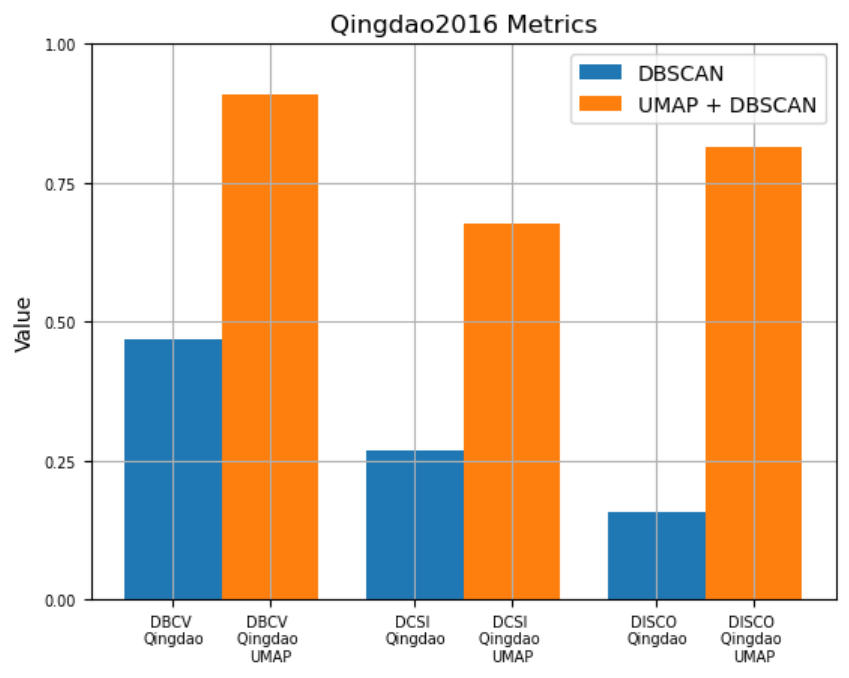}
    (c)~\includegraphics[width=0.45\linewidth]{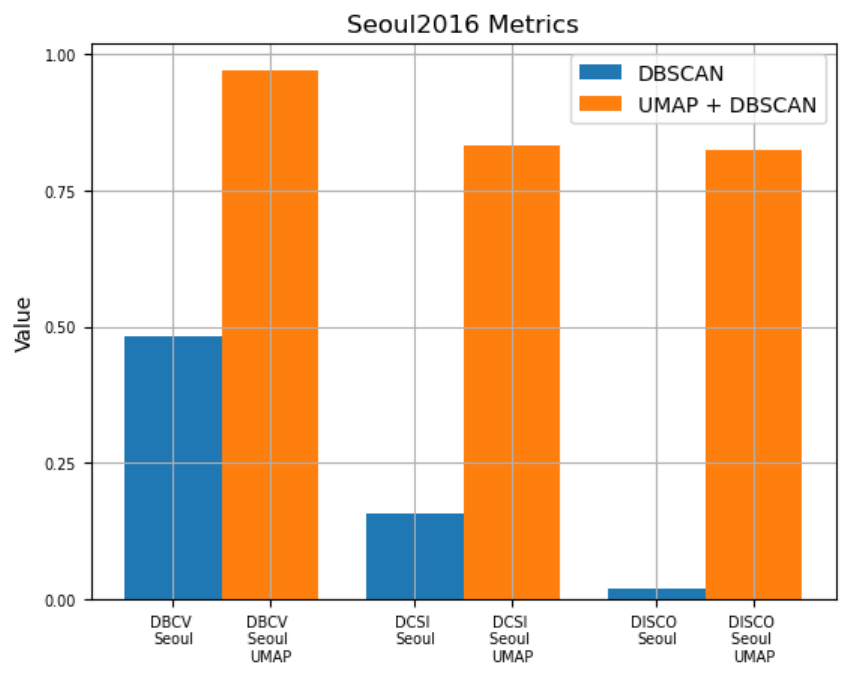}
    \caption{\textbf{Clustering results on the three datasets}.
    Barcharts of the clustering results measured through DBCV, DCSI, and DISCO on the three datasets: Heidelberg2019 (a), Qingdao2016 (b), and Seoul2016 (c).
    The list of hyperparameters' values follows.
    (a)~Heidelberg2019  DBSCAN: epsilon 150 and minimal points 3.
    Heidelberg2019 UMAP + DBSCAN: DBSCAN epsilon 0.7 and minimal points 13; UMAP number of neighbors 22 and minimum distance 0.05.
    (b)~Qingdao2016 DBSCAN: epsilon 7 and minimal points 4.
    Qingdao2016 UMAP + DBSCAN: DBSCAN epsilon 0.7 and minimal points 3; UMAP number of neighbors 9 and minimum distance 0.01.
    (c)~Seoul2016 DBSCAN: epsilon 61 and minimal points 13.
    Seoul2016 UMAP + DBSCAN:  DBSCAN epsilon 0.8 and minimal points 5; UMAP number of neighbors 10 and minimum distance 0.1.
    For UMAP, the number of components is always 2.
\label{fig:BARCHARTS-CLUSTERING-RESULTS}}
 \end{center}
\end{figure}

We report the results of our experiments in \autoref{tab:RESULTS-DETAILS} and \autoref{fig:BARCHARTS-CLUSTERING-RESULTS}.
Using UMAP with appropriately chosen hyperparameters improved clustering performance on all three datasets across all three evaluation metrics, while satisfying the three constraints we imposed.

\begin{table}[httb!] \footnotesize
     \centering
    \label{tab:Heidelberg2019-RESULTS-DETAILS}
    \begin{tabularx}{0.75\textwidth}{l c c}
    \toprule
    & \textbf{DBSCAN} & \textbf{UMAP + DBSCAN} \\
    \midrule
    \rowcolor{LightBlue}  & \textbf{Heidelberg2019} & \textbf{Heidelberg2019} \\
    \multirow{2}{*}{clusters' elements} & {cl~1: 243; cl~2: 102; cl~3: 7;} & \multirow{2}{*}{cl~1: 397; cl~2: 59} \\
     & {cl~4: 3; cl~5: 4; cl~6: 5} &  \\
     \rowcolor{LightBlue} noise points & 92 & 0 \\
     DBCV & 0.566 & 0.945 \\
     \rowcolor{LightBlue} DCSI & 0.217 & 0.730 \\
     DISCO & 0.297 & 0.801 \\
     \rowcolor{LightBlue} clusters' dimension \(>\) 10\% & \textcolor{red}{\ding{55}} & \textcolor{teal}{\checkmark} \\
     noise points \(<\) 20\%  & \textcolor{red}{\ding{55}} & \textcolor{teal}{\checkmark} \\
     \rowcolor{LightBlue} DBCV \(>\) 0.5 & \textcolor{teal}{\checkmark} & \textcolor{teal}{\checkmark} \\
    \midrule
     & \textbf{Qingdao2016} & \textbf{Qingdao2016} \\
     \rowcolor{LightBlue} clusters' elements & {cl~1: 181; cl~2: 4} & {cl~1: 124; cl~2: 76} \\
     noise points & 15 & 0 \\
     \rowcolor{LightBlue} DBCV & 0.467 & 0.909 \\
     DCSI & 0.269 & 0.676 \\
     \rowcolor{LightBlue} DISCO & 0.158 & 0.814 \\
     clusters' dimension \(>\) 10\% & \textcolor{red}{\ding{55}} & \textcolor{teal}{\checkmark} \\
     \rowcolor{LightBlue} noise points \(<\) 20\%  & \textcolor{teal}{\checkmark} & \textcolor{teal}{\checkmark} \\
     DBCV \(>\) 0.5 & \textcolor{red}{\ding{55}} & \textcolor{teal}{\checkmark} \\
    \midrule
     \rowcolor{LightBlue}  & \textbf{Seoul2016} & \textbf{Seoul2016} \\
     clusters' elements & {cl~1: 15; cl~2: 74} & {cl~1: 83; cl~2: 21} \\
     \rowcolor{LightBlue} noise points & 15 & 0 \\
     DBCV & 0.482 & 0.969 \\
     \rowcolor{LightBlue} DCSI & 0.158 & 0.831 \\
     DISCO & 0.020 & 0.825 \\
     \rowcolor{LightBlue} clusters' dimension \(>\) 10\% & \textcolor{teal}{\checkmark} & \textcolor{teal}{\checkmark} \\
     noise points \(<\) 20\%  & \textcolor{teal}{\checkmark} & \textcolor{teal}{\checkmark} \\
     \rowcolor{LightBlue} DBCV \(>\) 0.5 & \textcolor{red}{\ding{55}} & \textcolor{teal}{\checkmark} \\
    \bottomrule
    \end{tabularx}
    \caption{\textbf{Details of the clustering results on the three datasets.}
    cl X: number of elements of cluster X. For example, ``cl 1: 243'' means that the first cluster contains 243 elements. 
    \textcolor{red}{\ding{55}}: unmet condition.
    \textcolor{teal}{\checkmark}: met condition.
    DBCV interval: $[-1,+1]$.
    DCSI interval: $[0,1]$.
    DISCO interval: $[-1,+1]$.
    Interpretation of the values of DBCV, DCSI, and DISCO: the higher, the better.
    \label{tab:RESULTS-DETAILS}}
\end{table}

For the Heidelberg2019 dataset, DBSCAN alone identified six clusters, some of which had sizes below 10\%, along with a high number of noise points. 
In contrast, the combination of UMAP and DBSCAN identified two clusters, achieving DBCV = 0.945, DCSI = 0.73, and DISCO = 0.801.
For the Qingdao2016 dataset, DBSCAN alone produced two clusters, one of which was smaller than 10\%, with a sufficiently low number of noise points. 
In this case, UMAP followed by DBSCAN also identified two clusters, achieving an almost perfect DBCV score (0.909) and higher DCSI and DISCO values compared to DBSCAN alone.
For the Seoul2016 dataset, all three constraints were satisfied by both DBSCAN alone and the UMAP~+~DBSCAN approach. 
However, the latter method substantially improved the evaluation metrics: DBCV increased from 0.482 to 0.969, DCSI from 0.158 to 0.831, and DISCO from 0.02 to 0.825.

\section{Conclusion}
\label{sec:CONCLUSIONS}

Our results confirm that clustering preceded by dimensionality reduction is more effective than clustering alone when applied to electronic health records of patients with mammary carcinoma. The clustering analysis yielded consistent results across all three independent datasets and all three evaluation metrics, suggesting that the proposed approach has a degree of generalizability.
Regarding the limitations of this study, we acknowledge that we tested only one clustering method (DBSCAN) and one dimensionality reduction technique (UMAP), without exploring alternatives.
In future work, we plan to evaluate additional density-based clustering methods, such as HDBSCAN and OPTICS, as well as other dimensionality reduction algorithms, including $t$-SNE, PaCMAP, and PHATE. Moreover, we aim to extend this work by providing a medical interpretation of the patients' clusters identified by DBSCAN across the three datasets.

\section*{Conflict of interests}
\label{sec:CONFLICT-OF-INTERESTS}
The authors declare they have no conflict of interest.

\section*{Acknowledgments}
\label{sec:ACKNOWLEDGMENTS}
The authors acknowledge the use of Ecosia AI Chat for English proof-reading and grammar correction of this article. 

\section*{Funding}
The work of D.C. is funded by the Italian Ministero Italiano delle Imprese e del Made in Italy under the Digital Intervention in Psychiatric and Psychologist Services~(DIPPS) (project code F/310240/01-04/X56) programme within the framework ‘‘Innovation Agreements’’ (Accordi per l’Innovazione) and is supported by Ministero dell’Università e della Ricerca of Italy under the ``Dipartimenti di Eccellenza 2023-2027'' ReGAInS grant assigned to Dipartimento di Informatica Sistemistica e Comunicazione at Università di Milano-Bicocca. 
The funders had no role in study design, data collection and analysis, decision to publish, or preparation of the manuscript.

\section*{Availability of data and software code}
\label{sec:AVAILABILITY}
The analyzed datasets are publicly available on Figshare under the CC BY 4.0 license at the following URLs:   \\
-- Heidelberg2019 dataset~\url{https://tinu.be/Heidelberg2019} \\
-- Qingdao2016 dataset~\url{https://tinu.be/Qingdao2016}  \\
-- Seoul2016 dataset~\url{https://tinu.be/Seoul2016}

Our Python software code is openly available at \url{https://github.com/n-benvenuto/breast-cancer-clustering}

\footnotesize
\bibliographystyle{unsrt}
\bibliography{cibb_bibliography_file} 

@inproceedings{allaoui2020considerably,
  title={Considerably improving clustering algorithms using {UMAP} dimensionality reduction technique: a comparative study},
  author={Allaoui, Mebarka and Kherfi, Mohammed Lamine and Cheriet, Abdelhakim},
  booktitle={Proceedings of ICISP~2020},
  pages={317--325},
  year={2020},
  note={{DOI~URL}:~\url{https://doi.org/10.1007/978-3-030-51935-3_34}},   
  organization={Springer}
}

@article{Healy2024,
  title = {Uniform manifold approximation and projection},
  volume = {4},
  ISSN = {2662-8449},
  note = {{DOI~URL}:~\url{http://doi.org/10.1038/s43586-024-00363-x}},
  number = {1},
  journal = {Nature Reviews Methods Primers},
  author = {Healy,  John and McInnes,  Leland},
  year = {2024}
}

@inproceedings{khan2014dbscan,
  title={{DBSCAN}: past, present and future},
  author={Khan, Kamran and Rehman, Saif Ur and Aziz, Kamran and Fong, Simon and Sarasvady, Sababady},
  booktitle={Proceedings of ICADIWT~2014 -- the 5th International Conference on the Applications of Digital Information and Web Technologies},
  pages={232--238},
  note = {{DOI~URL}:~\url{http://doi.org/10.1109/icadiwt.2014.6930596}},
  year={2014},
  organization={IEEE}
}

@inproceedings{moulavi2014density,
  title={Density-based clustering validation},
  author={Moulavi, Davoud and Jaskowiak, Pablo A and Campello, Ricardo JGB and Zimek, Arthur and Sander, J{\"o}rg},
  booktitle={Proceedings of ICDM 2014},
  pages={839--847},
  year={2014},
  note = {{DOI~URL}:~\url{https://doi.org/10.1137/1.9781611973440.96}},
  organization={SIAM}
}

@article{gauss2023dcsi,
  title={{DCSI}--An improved measure of cluster separability based on separation and connectedness},
  author={Gauss, Jana and Scheipl, Fabian and Herrmann, Moritz},
  journal={arXiv preprint arXiv:2310.12806},
  note = {{DOI~URL}:~\url{https://arxiv.org/abs/2310.12806}},
  year={2023}
}

@article{beer2025disco,
  title={Internal Evaluation of Density-Based Clusterings with Noise},
  author={Beer, Anna and Krieger, Lena and Weber, Pascal and Ritzert, Martin and Assent, Ira and Plant, Claudia},
  note = {{DOI~URL}:~\url{https://arxiv.org/abs/2503.00127}},
  journal={arXiv preprint arXiv:2503.00127},
  year={2025}
}

@article{lee2016clinicopathological,
  title = {Clinicopathological features and prognostic factors affecting survival outcomes in isolated locoregional recurrence of breast cancer: single-institutional series},
  author = {Lee, Min-Young and Chang, Won Jin and Kim, Hae Su and Lee, Ji Yun and Lim, Sung Hee and Lee, Jeong Eon and Kim, Seok Won and Nam, Seok Jin and Ahn, Jin Seok and Im, Young-Hyuck and Park, Yeon Hee },
  journal = {PLOS One},
  volume = {11},
  number = {9},
  pages = {e0163254},
  year = {2016},
  note = {{DOI~URL}:~\url{https://doi.org/10.1371/journal.pone.0163254}},
  publisher = {Public Library of Science San Francisco, CA USA}
}

@article{jiao2016elevated,
  title = {Elevated serum levels of retinol-binding protein 4 are associated with breast cancer risk: a case-control study},
  author = {Jiao, Congcong and Cui, Lianhua and Ma, Aiguo and Li, Na and Si, Hongzong},
  journal = {PLOS One},
  volume = {11},
  number = {12},
  pages = {e0167498},
  year = {2016},
  note = {{DOI~URL}:~\url{https://doi.org/10.1371/journal.pone.0167498}},
  publisher = {Public Library of Science San Francisco, CA USA}
}

@article{michel2019prediction,
  title = {Prediction of local recurrence risk after neoadjuvant chemotherapy in patients with primary breast cancer: clinical utility of the {MD Anderson Prognostic Index}},
  author = {Michel,  Laura L. and Sommer,  Laura and González Silos,  Rosa and Lorenzo Bermejo,  Justo and von Au,  Alexandra and Seitz,  Julia and Hennigs,  André and Smetanay,  Katharina and Golatta,  Michael and Heil,  J\"{o}rg and Sch\"{u}tz,  Florian and Sohn,  Christof and Schneeweiss,  Andreas and Marmé,  Frederik},
  journal = {PLOS One},
  volume = {14},
  number = {1},
  pages = {e0211337},
  year = {2019},
  note = {{DOI~URL}:~\url{https://doi.org/10.1371/journal.pone.0211337}},
  publisher = {Public Library of Science San Francisco, CA USA}
}

@article{chen2005breast,
  title = {Breast conservation after neoadjuvant chemotherapy: a prognostic index for clinical decision-making},
  author = {Chen,  Allen M. and Meric‐Bernstam,  Funda and Hunt,  Kelly K. and Thames,  Howard D. and Outlaw,  Elesyia D. and Strom,  Eric A. and McNeese,  Marsha D. and Kuerer,  Henry M. and Ross,  Merrick I. and Singletary,  S. Eva and Ames,  Fredrick C. and Feig,  Barry W. and Sahin,  Aysegul A. and Perkins,  George H. and Babiera,  Gildy and Hortobagyi,  Gabriel N. and Buchholz,  Thomas A.},
  journal = {Cancer},
  volume = {103},
  number = {4},
  pages = {689--695},
  year = {2005},
  note = {{DOI~URL}:~\url{https://doi.org/10.1002/cncr.20815}},
  publisher = {Wiley Online Library}
}
\normalsize

\end{document}